\documentclass[journal,twoside,web]{ieeecolor}
\usepackage{generic}
\usepackage{cite}
\usepackage{amsmath,amssymb,amsfonts}
\usepackage{graphicx}
\usepackage{textcomp}
\usepackage{multirow}
\usepackage{booktabs}
\usepackage{threeparttable}
\usepackage{hyperref}

\def\BibTeX{{\rm B\kern-.05em{\sc i\kern-.025em b}\kern-.08em
    T\kern-.1667em\lower.7ex\hbox{E}\kern-.125emX}}
\begin{document}
\title{Two-Stage Personalized Gait Phase Estimation in Stroke Survivors During Exoskeleton-Assisted Walking: An Offline Feasibility Study}
\author{Hyungseok Ryu, \IEEEmembership{Student Member, IEEE}, and Pilwon Hur, \IEEEmembership{Member, IEEE}
\thanks{}
\thanks{H. Ryu and P. Hur are with the Department of Mechanical and Robotics Engineering, Gwangju Institute of Science and Technology (GIST), Gwangju 61005, Republic of Korea (e-mail: guswls1287@gm.gist.ac.kr; pilwonhur@gist.ac.kr). }
}

\maketitle

\begin{abstract}
This study evaluated personalized gait phase estimation for stroke survivors using functional inertial measurement unit (IMU) alignment and two-stage sequential adaptation of models pre-trained on healthy gait. The estimator used signals from a thigh-mounted IMU. Heel force-sensitive resistor measurements provided reference phase labels for offline adaptation and evaluation. Stage~1 established a distillation-regularized participant-specific model, and Stage~2 performed conditional refinement using low-rank adaptation. Long Short-Term Memory (LSTM), Temporal Convolutional Network (TCN), and Transformer models were evaluated in five stroke survivors walking with a powered knee exoskeleton using leave-one-subject-out hyperparameter selection and sequential test-then-adapt Stage~2 replay. Relative to the non-adapted baselines, Stage~1+2 reduced the mean participant-wise phase root mean square error by 84.2\%, 77.0\%, and 60.7\%, respectively. The Transformer achieved the lowest final error ($2.90 \pm 1.13$\% of the gait cycle) and heel-strike timing error ($23.7 \pm 4.5$~ms). Policy-specific ablations showed that every-cycle updates generally produced the lowest or near-lowest error, whereas conditional updating reduced the update frequency with small accuracy differences. After personalization, alignment produced model-dependent changes in phase error while preserving or improving heel-strike detection and reducing heel-strike timing error for the LSTM and Transformer. Concurrent embedded tests showed that the TCN and Transformer maintained 100-Hz inference during Stage~2 updates without deadline misses, whereas the LSTM missed the 10-ms deadline in 6.6\% of inferences. All updates completed within 0.8~s. These results support the offline feasibility and embedded computational timing of the proposed framework for exoskeleton-assisted walking.
\end{abstract}



\section{Introduction}
\label{sec:introduction}

Stroke-related motor impairments commonly compromise independent walking through spatiotemporal asymmetry and increased energy expenditure~\cite{olney1996hemiparetic, awad2015walking, finley2017associations}. Restoring efficient gait is therefore central to functional independence and quality of life after stroke~\cite{perry1995classification, lord2004community, kramer2016energy}. Lower-limb robotic exoskeletons can provide intensive, task-specific gait assistance~\cite{nolan2020robotic, mehrholz2025electromechanical}, but effective synchronization requires accurate continuous gait-phase estimation~\cite{kang2019real, lee2021continuous, hong2022piecewise}, which remains challenging for neurologically impaired gait~\cite{azevedo2014continuous, shushtari2022ultra}.

Post-stroke gait exhibits substantial between- and within-participant variability due to differences in impairment severity, compensatory strategies, and recovery progression~\cite{beyaert2015gait, calafiore2021efficacy}. This variability can limit the generalization of gait phase estimation models pre-trained on large-scale healthy gait datasets to paretic gait dynamics. Performance degradation may therefore occur when these models are applied under clinical walking conditions that differ from those represented during pre-training~\cite{sanchez2018individual}.

For exoskeleton-assisted walking, wearable sensing should minimize setup and calibration burden. A single thigh-mounted inertial measurement unit (IMU) offers a practical balance between hardware simplicity and signal reliability because thigh kinematics remain relatively consistent despite distal impairments such as foot drop~\cite{choi2023walking}. Accurate and adaptive estimation within this sensor constraint is important for phase-scheduled assistance.

Beyond gait variability and sensor minimalism, gait phase estimation models pre-trained on open-source datasets face additional challenges related to coordinate consistency. Variations in sensor orientation, mounting location, limb side (left versus right), and coordinate frame definitions across users and devices introduce spatial mismatches that distort inertial signal patterns and degrade model performance~\cite{banos2014dealing, tan2019influence}. When combined with participant-specific gait characteristics after stroke, these inconsistencies further increase the mismatch between pre-training data and paretic gait data~\cite{choi2021unsupervised, kang2025online}. Mitigating this compounded mismatch motivates adaptation strategies that account for both sensor alignment and personalized gait dynamics without extensive participant-specific data collection.

Data-driven approaches can estimate continuous gait phase directly from inertial measurements~\cite{medrano2023real, ji2025human}. A recent framework combined temporal convolution and Transformer-based fusion across pelvis and bilateral-leg IMUs, with limited participant-specific fine-tuning to reduce between-participant variability~\cite{ji2025human}. Such offline fine-tuning provides initial calibration but does not refine the model as gait changes across later trials or assistance conditions.

An online adaptation framework for stroke survivors improved phase-estimation accuracy and joint-torque profiles relative to a fixed model during exoskeleton-assisted walking~\cite{kang2025online}. However, a framework that explicitly separates initial participant calibration from low-rank within-session refinement remains underexplored.

To address this gap, we propose a two-stage personalization framework that combines participant-specific calibration with cycle-wise refinement using Low-Rank Adaptation (LoRA). The estimator uses a single thigh-mounted IMU as its inference input. The framework was evaluated offline using phase labels derived from a heel force-sensitive resistor (FSR) and overground walking data from stroke survivors under multiple exoskeleton assistance conditions. Concurrent inference and adaptation timing was also assessed on an embedded platform.

The contributions of this study are summarized as follows.
First, we propose a personalization framework that combines functional IMU alignment and participant-specific calibration with parameter-efficient within-session refinement.
Second, we evaluate alternative Stage~1 calibration objectives and compare conditional, every-cycle, and fixed-period Stage~2 update policies across three model architectures.
Third, we evaluate the personalized estimator under multiple exoskeleton-assisted walking conditions and assess the computational timing of concurrent inference and adaptation on an embedded platform.

\section{Methods}


\begin{figure*}[t!]
    \centering
    \includegraphics[width=0.95\linewidth]{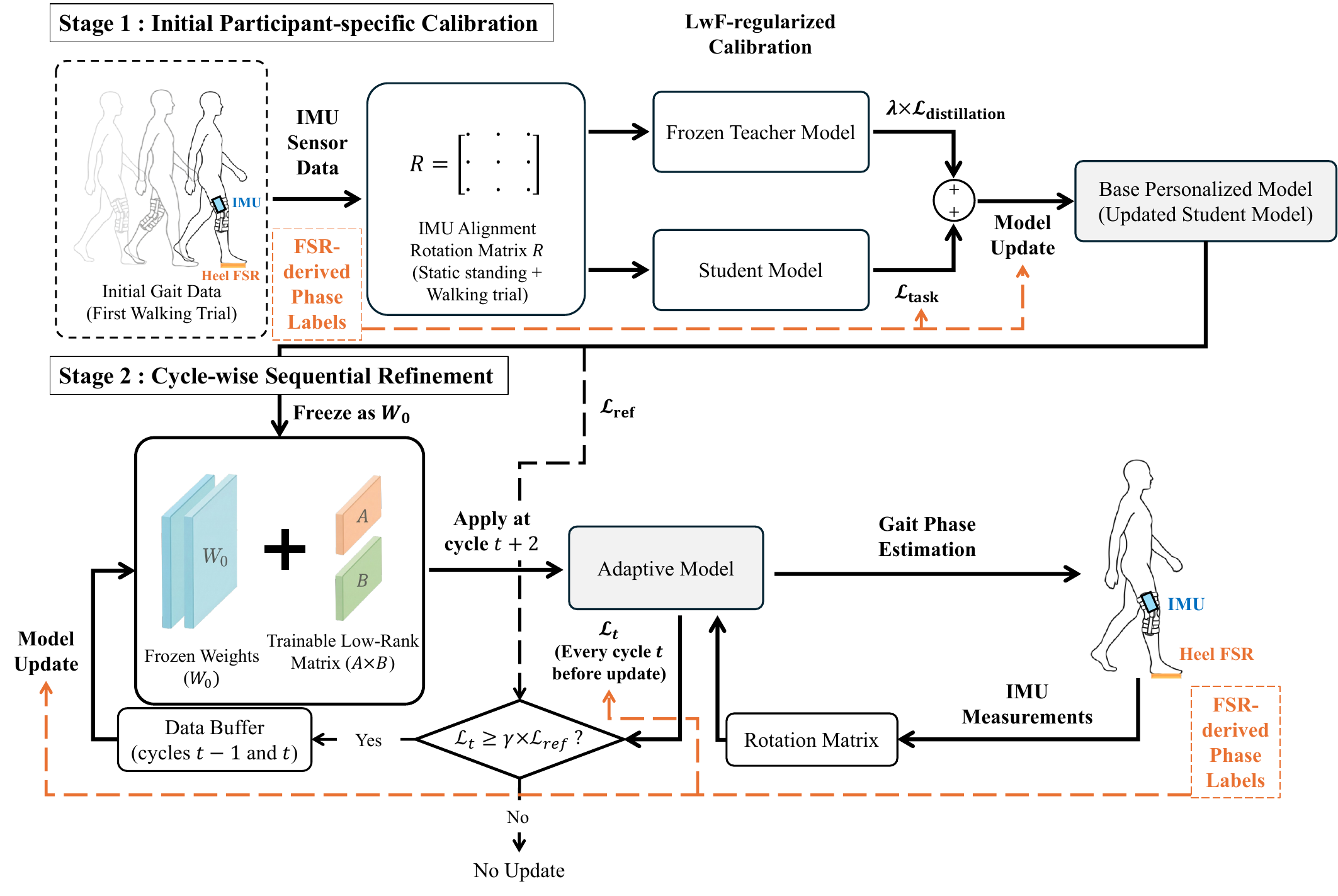}
    \caption{Structure of the proposed two-stage personalization framework. Stage~1 calibrates the pre-trained model using aligned thigh-IMU signals and heel-FSR-derived phase labels. Stage~2 performs cycle-wise LoRA refinement during sequential replay, with updates triggered by the phase loss relative to the Stage~1 reference loss.}
    \label{fig:adaptation_strategy}
\end{figure*}

\subsection{Open-source Datasets for Pre-training}
\label{sec:dataset}
Two publicly accessible IMU-based gait datasets were used to pre-train baseline models~\cite{wiles2023nonan, wiles2025nonan}. The datasets included full-body IMU recordings from healthy young and older adults walking at self-selected speeds along a 200-meter indoor track. Noraxon Ultium Motion sensors provided tri-axial acceleration, tri-axial angular velocity, and segment orientation at 200~Hz across sixteen anatomical locations. Heel-strike (HS) events were supplied by the manufacturer's event detection algorithm for cycle segmentation. A total of 76 participants were included.

For computational efficiency, a fixed subset of twenty-five participants was selected from the full datasets. Fifteen participants were used for training, five for validation, and five for testing. Only straight-path walking sequences were retained to exclude turning variability. To match the target sensor configuration, IMU signals from the right thigh were used. These signals were resampled from 200~Hz to 100~Hz and processed with a fourth-order Butterworth low-pass filter with a 10~Hz cutoff.

Reference gait phase was generated from the supplied HS events. A gait cycle was defined as the interval between consecutive right-leg HS events at times $t_k$ and $t_{k+1}$. Within this interval ($t_k \leq t < t_{k+1}$), normalized gait progression was defined as $\tau(t) = (t-t_k)/(t_{k+1}-t_k)$ and mapped to the phase angle $\phi(t)=2\pi\tau(t)$. To avoid the discontinuity from $2\pi$ to $0$, the phase was encoded as a continuous unit vector:
\begin{equation}
\label{eq:phase}
\phi_{\sin}(t) = \sin(\phi(t)), \quad \phi_{\cos}(t) = \cos(\phi(t)).
\end{equation}
Additionally, the phase rate is defined as the inverse of the gait cycle duration to represent phase dynamics:
\begin{equation}
\label{eq:phase_dot}
\dot{\tau}(t) = \frac{1}{t_{k+1} - t_k}.
\end{equation}
This formulation enables the model to jointly learn gait phase representation and its temporal dynamics. Consequently, all models were trained to regress the three-dimensional target vector $\mathbf{y} = [\phi_{\sin}, \phi_{\cos}, \dot{\tau}]^\top$ at each time step.

\subsection{Baseline Gait Phase Estimation Models}
\label{sec:model}
To evaluate the proposed personalization framework across complementary temporal modeling approaches, we implemented a Long Short-Term Memory (LSTM) network, a Temporal Convolutional Network (TCN), and an encoder-only Transformer. The LSTM models the quasi-periodic gait sequence through recurrent memory~\cite{hochreiter1997long,lee2021continuous}, the TCN supports efficient causal phase tracking through hierarchical dilated convolutions~\cite{bai2018empirical,guo2023speed}, and the Transformer captures complex temporal dependencies through multi-head self-attention~\cite{vaswani2017attention,ji2025human}.

During inference, all models rely exclusively on a single thigh-mounted IMU. Each model receives a 120-sample input window containing seven channels: tri-axial acceleration, tri-axial angular velocity, and thigh pitch angle. This window represents 1.2~s at 100~Hz and advances by one sample at each step. The heel-FSR signal is not included in the input window or provided to any gait phase estimation model.

All architectures were trained to regress the same three-dimensional target vector $\mathbf{y} = [\phi_{\sin}, \phi_{\cos}, \dot{\tau}]^\top$. Let $\mathcal{L}_{\mathrm{phase}}$ denote the mean squared error (MSE) of the sine and cosine components and $\mathcal{L}_{\mathrm{rate}}$ the MSE of $\dot{\tau}$. The task loss used for model parameter updates was
\begin{equation}
\mathcal{L}_{\mathrm{task}}=\mathcal{L}_{\mathrm{phase}}+0.02\mathcal{L}_{\mathrm{rate}}.
\end{equation}
The smaller phase-rate weight prevented this component from dominating optimization. Model validation used $\mathcal{L}_{\mathrm{phase}}$ only, reflecting its central role in continuous phase estimation. Architecture-specific hyperparameters were selected via grid search based on validation performance. Detailed architectural configurations and baseline performance on the open-source test set are reported in Tables~S1 and S2.

\subsection{Proposed Personalization Framework}

\subsubsection{IMU Axis Alignment}
\label{sec:IMU_alignment}

To reduce coordinate-frame differences between the open-source data and the experimental IMU, we constructed a functional orthonormal basis separately for the reference and target sensors. Let $\mathbf{g}$ denote the normalized mean acceleration vector during static standing and $\mathbf{p}$ the principal eigenvector obtained by applying principal component analysis (PCA) to the thigh angular velocity during walking~\cite{fry2021method, mcgrath2018auto}. The target principal-vector direction was standardized according to the participant's affected side. Each basis was constructed as
\begin{equation}
\mathbf{e}_x=\mathbf{g},\quad
\mathbf{e}_z=\frac{\mathbf{e}_x\times\mathbf{p}}
{\|\mathbf{e}_x\times\mathbf{p}\|},\quad
\mathbf{e}_y=\mathbf{e}_z\times\mathbf{e}_x,
\end{equation}
with $B=[\mathbf{e}_x\;\mathbf{e}_y\;\mathbf{e}_z]$. The target-to-reference rotation was then calculated as
\begin{equation}
R=B_{\mathrm{ref}}B_{\mathrm{target}}^{\top},
\end{equation}
and applied to both the acceleration and angular velocity signals. This procedure aligns sensor coordinates without directly matching participant-specific gait waveforms.

\subsubsection{Two-stage Adaptation Strategy}

The proposed framework separates participant-specific calibration from lightweight sequential refinement (Fig.~\ref{fig:adaptation_strategy}). Stage~1 establishes a calibrated participant model from the first walking trial. In the primary configuration, this calibration uses Learning without Forgetting (LwF) as a regularized objective, while standard fine-tuning is evaluated as an alternative~\cite{li2017learning}. Stage~2 constitutes the post-calibration sequential component of the framework and uses LoRA to update a limited set of parameters as gait cycles are encountered~\cite{hu2022lora}. Stage~1 calibration and triggered Stage~2 parameter updates use $\mathcal{L}_{\mathrm{task}}$, whereas the Stage~2 trigger uses the phase-only losses $\mathcal{L}_{ref}$ and $\mathcal{L}_t$ defined below. In the present study, both stages were conducted offline using recorded walking data and phase targets generated from heel-FSR-derived HS events; Stage~2 was evaluated through sequential replay of the recorded gait cycles.

\noindent \quad \textit{\textbf{(1) Stage 1 -- Initial Participant-Specific Calibration:}} In Stage~1, the pre-trained model is calibrated offline for each participant using the first walking trial, which contained four to seven retained gait cycles across participants. Because calibration was limited to this single short trial, LwF was used as a conservative regularizer to constrain large deviations from the pre-trained model while optimizing the FSR-supervised task objective. The cycle boundaries and task targets are generated from heel-FSR-derived HS events. Prior to calibration, the functional axis alignment matrix $R$ is computed using a short static standing trial and the same walking trial to ensure spatial consistency. The resulting matrix is then stored for subsequent use. For the primary LwF-based implementation, the pre-trained model serves as a frozen teacher and the adapting model as a student. Let $\hat{\boldsymbol{\phi}}=[\hat{\phi}_{\sin},\hat{\phi}_{\cos}]^\top$. The distillation loss is defined as
\begin{equation}
\mathcal{L}_{\mathrm{distill}}
=\operatorname{MSE}(\hat{\boldsymbol{\phi}}^{S},\hat{\boldsymbol{\phi}}^{T})
+0.02\operatorname{MSE}(\hat{\dot{\tau}}^{S},\hat{\dot{\tau}}^{T}),
\end{equation}
where superscripts $S$ and $T$ denote the student and teacher outputs, respectively. The Stage~1 objective is
\begin{equation}
\mathcal{L}_{\mathrm{S1}}
=\mathcal{L}_{\mathrm{task}}+\lambda\mathcal{L}_{\mathrm{distill}},
\end{equation}
where the FSR-derived phase targets provide $\mathcal{L}_{\mathrm{task}}$ and $\lambda$ controls the distillation regularization.

Stage~1 training uses all complete gait cycles from the first walking trial. The learning rate and number of training epochs are selected from the training-side participants in each leave-one-subject-out (LOSO) fold. The selected configuration is then fixed and applied to the held-out participant without internal validation or early stopping. After training, the adapted model is re-evaluated over the calibration cycles. The reference loss $\mathcal{L}_{ref}$ is the mean squared error between the predicted and FSR-derived sine and cosine phase components; it excludes the distillation and phase-rate terms.

The first walking trial is used only for functional alignment and Stage~1 calibration. It is excluded from the performance comparisons of the non-adapted baseline, Stage~1-only, and Stage~1+2 models, which use the same gait cycles from all subsequent trials. Stage~2 begins at the first gait cycle of the next trial without an additional performance threshold or manual decision.

\noindent \quad \textit{\textbf{(2) Stage 2 -- Cycle-Wise Sequential Refinement via LoRA:}} After calibration, Stage~2 sequentially refines the model as gait cycles are encountered. Incoming IMU measurements are transformed using the alignment matrix $R$ obtained in Stage~1. To limit the number of trainable parameters and reduce the computational cost of each update, LoRA is employed by freezing the weights from Stage~1 and injecting trainable low-rank decomposition matrices into selected linear modules~\cite{hu2022lora}.

Target layers were selected to prioritize high-level task-specific features: the final fully connected layers for the LSTM, the output linear projection for the TCN, and the output projections of self-attention and feed-forward modules for the Transformer \cite{dettmers2023qlora}. With the LOSO-selected rank of 16 used in the primary configuration, this strategy resulted in model-specific trainable parameter proportions of 0.52\% for the LSTM, 0.12\% for the TCN, and 18.81\% for the Transformer. Rather than enforcing a uniform update budget, this approach focuses adaptation on layers directly involved in phase prediction for each architecture.

Stage~2 operates cycle by cycle through offline sequential replay. Each gait cycle is segmented by consecutive heel-FSR-derived HS events. The corresponding phase targets are used both to compute the cycle-level trigger loss $\mathcal{L}_t$ and to supervise any triggered update. For gait cycle $t$, $\mathcal{L}_t$ is the mean sine--cosine phase MSE over its samples, using the same phase-only definition as $\mathcal{L}_{ref}$. An update is triggered when this loss exceeds a relative threshold:
\begin{equation}
\mathcal{L}_t \geq \gamma \, \mathcal{L}_{ref},
\end{equation}
where $\gamma$ controls the trigger sensitivity relative to the Stage~1 calibration error.
This criterion was chosen to make Stage~2 updates responsive to observed prediction deterioration rather than enforcing an update every cycle or at a fixed interval.
When triggered, the LoRA parameters are updated for one epoch by minimizing $\mathcal{L}_{\mathrm{task}}$ over the current cycle and a replay buffer containing the immediately preceding cycle. For the first Stage~2 cycle, the buffer is empty, so any triggered update uses only the current cycle; a two-cycle update set is used thereafter. The trigger therefore uses the phase-only $\mathcal{L}_t$, whereas the parameter update also includes the weighted phase-rate term. This compact update set limits computation and reduces overfitting to the most recent cycle.

To prevent parameter changes within a gait cycle and accommodate update latency, parameters updated from cycle $t$ are applied at the onset of cycle $t+2$. Inference therefore uses fixed parameters within each cycle.

\begin{table}[t]
\centering
\caption{Hyperparameter candidates evaluated within each LOSO fold. The LwF weight applies only to LwF-based calibration, and $\gamma$ only to conditional Stage~2 updates.}
\label{tab:loso_candidates}
\begin{tabular}{@{}llp{0.43\columnwidth}@{}}
\toprule
\textbf{Stage} & \textbf{Hyperparameter} & \textbf{Candidate values} \\
\midrule
\multirow{3}{*}{Stage~1}
& Learning rate & $\{5\!\times\!10^{-5},1\!\times\!10^{-4},2\!\times\!10^{-4}\}$ \\
& LwF weight $\lambda$ & $\{0.1,0.2,0.4,0.8\}$ \\
& Epoch count & $\{5,10,20,30,40,50,75\}$ \\
\midrule
\multirow{3}{*}{Stage~2}
& Learning rate & $\{5\!\times\!10^{-5},1\!\times\!10^{-4},2\!\times\!10^{-4}\}$ \\
& LoRA $(r,\alpha)$ & $\{(4,8),(8,16),(16,32)\}$ \\
& Trigger threshold $\gamma$ & $\{1,2,3,4\}$ \\
\bottomrule
\end{tabular}
\end{table}

Adaptation hyperparameters were selected sequentially using a LOSO protocol within the stroke cohort. The candidate values are summarized in Table~\ref{tab:loso_candidates}. For each held-out participant and model architecture, every Stage~1 candidate was trained on the first trial of the four training-side participants and scored by the mean Stage~1 root mean square error (RMSE) over their subsequent trials. The lowest-RMSE configuration was selected, with the lower epoch count preferred in a tie.

After fixing the Stage~1 configuration, Stage~2 candidates were evaluated on the same four participants. Selection minimized their mean prequential phase RMSE, with mean update count used only to break a tie. The resulting configuration was applied to the held-out participant without further tuning.

For each architecture and held-out participant, a separate copy of the common pre-trained model was calibrated and sequentially adapted using only that participant's adaptation data. Adapted weights were not shared across participants. Stage~2 used a prequential test-then-adapt protocol over the subsequent trials: each cycle was evaluated before an update based on that cycle could affect later predictions~\cite{gama2013evaluating}. The recorded condition order was followed without resetting the model, optimizer, or replay buffer at condition transitions. The LOSO-selected hyperparameters for the primary method are detailed in Table~S3.

\begin{figure}[t]
    \centering
    \includegraphics[width=\columnwidth]{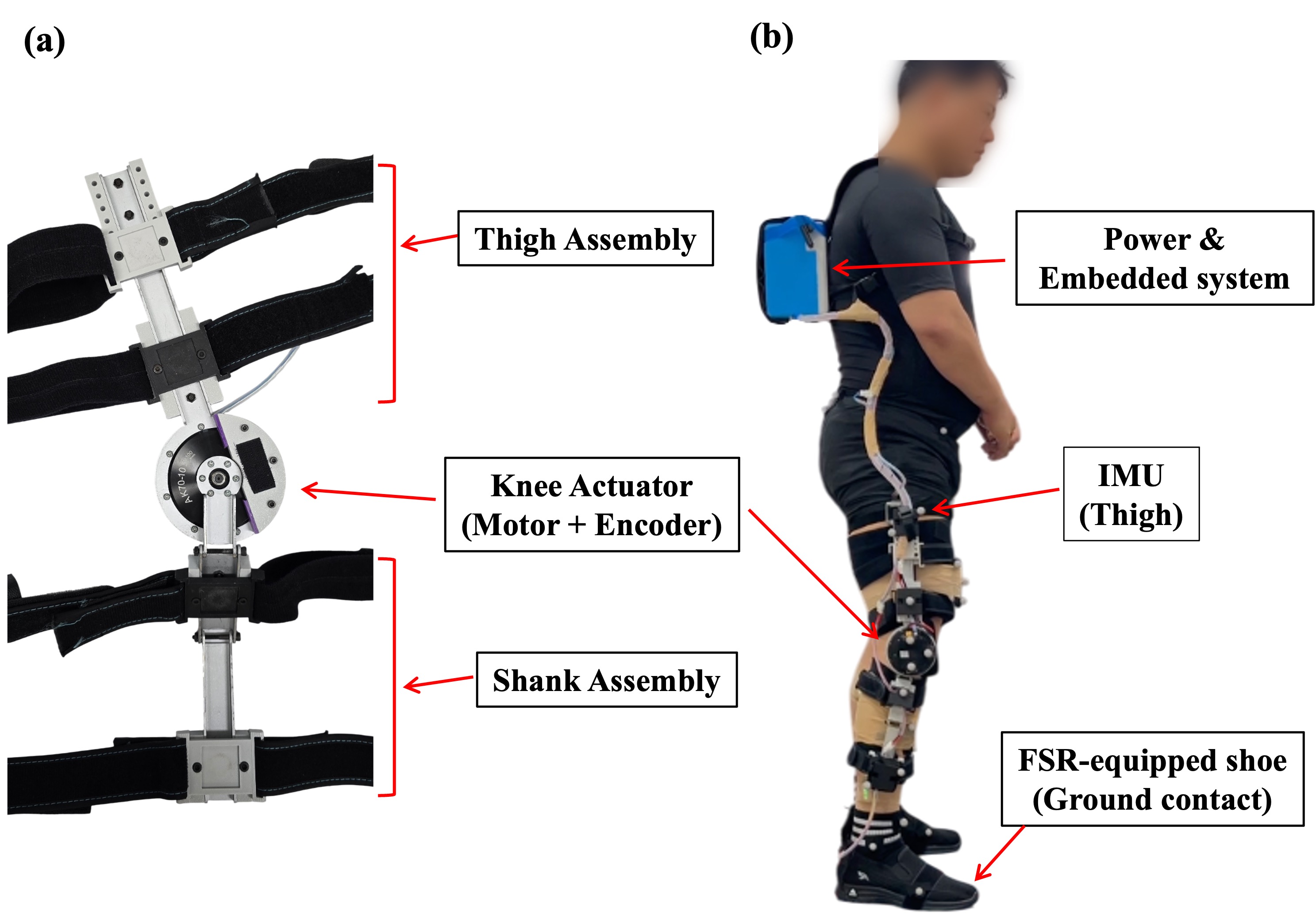}
    \caption{Experimental setup for overground walking experiments with stroke survivors.
    (a) Custom-built powered knee exoskeleton designed to assist knee movement during locomotion.
    (b) Sensor placement and experimental configuration during overground walking trials.
    }
    \label{fig:knee_exoskeleton}
\end{figure}

\subsection{Experimental Validation Setup}

\subsubsection{Powered Knee Exoskeleton System}
\label{sec:system}

\begin{table*}[t]
    \centering
    \begin{threeparttable}
    \caption{\textsc{Participant Demographics and Clinical Characteristics}}
    \label{tab:participant_info}
    \begin{tabular}{c c c c c c c c c c c}
    
    \hline
    \noalign{\vskip 3pt}
    
    \multirow{2}{*}{{Participant}} &
    \multirow{2}{*}{{Sex}} &
    \multicolumn{1}{c}{{Age}} &
    \multicolumn{1}{c}{{Height}} &
    \multicolumn{1}{c}{{Weight}} &
    \multicolumn{1}{c}{{Years since}} &
    \multirow{2}{*}{{Affected Side}} &
    \multirow{2}{*}{{Stroke Type}} &
    \multirow{2}{*}{{FAC}} &
    \multirow{2}{*}{Control Condition Order} \\
    & &
    \multicolumn{1}{c}{(yrs)} &
    \multicolumn{1}{c}{(cm)} &
    \multicolumn{1}{c}{(kg)} &
    \multicolumn{1}{c}{Stroke} &
    & & \\
    
    \noalign{\vskip 3pt}
    \hline
    \noalign{\vskip 2pt}
    S01 & M & 52 & 172 & 94 & 4 & Left & Ischemic & 3 & NA $\rightarrow$ PD $\rightarrow$ AIF \\
    S02 & M & 59 & 160 & 65 & 38 & Right & Ischemic & 4 & NA $\rightarrow$ AIF $\rightarrow$ PD \\
    S03 & M & 44 & 167 & 84 & 21 & Left & Ischemic & 4 & AIF $\rightarrow$ NA $\rightarrow$ PD \\
    S04 & M & 35 & 179 & 84 & 17 & Left & Ischemic & 4 & PD $\rightarrow$ AIF $\rightarrow$ NA\\
    S05 & M & 45 & 170 & 87 & 30 & Left & Ischemic & 4 & PD $\rightarrow$ AIF $\rightarrow$ NA \\
    \noalign{\vskip 2pt}
    \hline
    \end{tabular}
    
    \begin{tablenotes}[flushleft]
    \footnotesize
    \item Functional Ambulation Category (FAC), no assistance (NA), Proportional-Derivative (PD), and Active-Inference (AIF).
    \end{tablenotes}
    
    \end{threeparttable}
\end{table*}

A custom-built powered knee exoskeleton was used in the walking experiments (Fig.~\ref{fig:knee_exoskeleton}(a)). The device was worn on the participant’s affected side and provided assistive knee flexion–extension via a single actively driven joint actuated by a quasi-direct drive motor (AK70-10, T-Motor, Nanchang, China). Knee joint angle and angular velocity were measured using a high-resolution incremental encoder integrated within the actuator. A thigh-mounted IMU (ICM-20948, InvenSense, San Jose, CA, USA) provided tri-axial acceleration, tri-axial angular velocity, and the thigh pitch angle in the sagittal plane, which served as the input features for the gait phase estimator. Additionally, a heel-mounted FSR (FlexiForce A301, Tekscan, South Boston, MA, USA) was used to capture reference gait events. The FSR signal was not used as an input to the estimator; it was used to generate supervised phase labels for the offline Stage~1 and Stage~2 adaptation procedures and for final performance evaluation.

The exoskeleton operated under a two-level embedded architecture. Low-level current and position control was executed at 1000~Hz on a microcontroller (STM32 Nucleo, STMicroelectronics, Geneva, Switzerland). IMU and FSR acquisition and data logging were executed at 100~Hz on an onboard computer (Jetson Orin NX, NVIDIA, Santa Clara, CA, USA).

The powered knee exoskeleton was operated under three control conditions to generate diverse gait patterns: (1) no assistance (NA), with zero applied torque; (2) Proportional-Derivative (PD) control, which tracked a predefined trajectory; and (3) Active-Inference (AIF) control, which provided adaptive assist-as-needed torque. These conditions were used to capture varied interaction dynamics rather than to compare the controllers. The estimator did not influence exoskeleton control, and all personalized phase estimations were performed offline.

\subsubsection{Experimental Protocol}

Five individuals with stroke participated in the gait experiment. Their anthropometric and clinical characteristics are summarized in Table~\ref{tab:participant_info}, and the sensor placement is shown in Fig.~\ref{fig:knee_exoskeleton}(b). Participants were required to walk overground on a level surface without personal aids (e.g., canes or walkers) and to meet the requirements for safe powered-exoskeleton use. The protocol was approved by the Institutional Review Board at Gwangju Institute of Science and Technology (20250618-HR-83-08-04), and all participants provided written informed consent.

Each participant completed walking trials under the NA, PD, and AIF conditions. Their order was randomized for each participant. Before data collection, controller parameters were tuned for each condition, and participants practiced with the corresponding controller until stable gait was achieved.

For each condition, participants performed five overground walking trials along a straight 7~m walkway at a self-selected comfortable speed. Two researchers walked alongside each participant during all trials to ensure safety. Participants were allowed to rest between trials, and any trial was immediately terminated in the event of discomfort, fatigue, or equipment malfunction. IMU, FSR, and control logs were recorded synchronously at the configured system sampling rate. Due to a data-logging failure, one recording from S04 in the NA condition (trial 1) was unavailable and was therefore excluded from all analyses. To extract steady-state gait segments, the first one to two gait cycles of each trial and the final one to two cycles near trial termination were excluded to remove transient acceleration and deceleration effects. Consequently, a minimum of 57 steady-state gait cycles per participant were retained across the three conditions for subsequent offline analysis.

\subsubsection{Concurrent Inference and Adaptation Timing Evaluation}

The computational feasibility of concurrent inference and Stage~2 adaptation was evaluated on a Jetson Orin NX by replaying the recorded data. Consistent with heterogeneous on-device adaptation designs~\cite{piechocki2026empowering}, fixed operations were executed with TensorRT while trainable LoRA operations remained in PyTorch. The Stage~2 procedure was replayed while inference ran continuously at 100~Hz on the same embedded graphics processor.

After an initialization warm-up, mean and 99th-percentile (p99) inference latencies were calculated for each participant and summarized across participants as the mean $\pm$ standard deviation. An inference exceeding the 10~ms sampling interval was counted as a deadline miss, and the pooled miss rate was reported for each model. Adaptation time was pooled across triggered updates and summarized by the mean $\pm$ standard deviation, p99, and maximum. Skipped updates and update backlog were also recorded.

\subsection{Data Analysis}
Reference phase labels for the experimental data were generated from the heel-mounted FSR signal. HS events were identified by thresholding and used to segment gait cycles. The linear gait phase was encoded as $\mathbf{y} = [\phi_{\sin}, \phi_{\cos}, \dot{\tau}]^\top$ according to Eqs.~\ref{eq:phase} and \ref{eq:phase_dot}. These labels provided the Stage~1 task and reference losses, the Stage~2 trigger and update losses, and the final evaluation reference.

To evaluate the proposed estimator, we analyzed continuous phase accuracy and discrete gait-event detection. At each evaluated sample, the reference phase $\tau$ and predicted phase $\hat{\tau}$ were reconstructed from their respective sine and cosine components using $\operatorname{atan2}$. The resulting angles were wrapped to $[0,2\pi)$ and divided by $2\pi$, yielding normalized phases in $[0,1)$. Because gait phase is circular, the error for sample $i$ in gait cycle $c$ was defined as the shorter distance around the unit circle,
\begin{equation}
d_{c,i}=\min\!\left(
\left|\hat{\tau}_{c,i}-\tau_{c,i}\right|,
1-\left|\hat{\tau}_{c,i}-\tau_{c,i}\right|
\right).
\end{equation}
The cycle-wise phase RMSE was then calculated as
\begin{equation}
\mathrm{RMSE}_c~(\%GC)
=100\sqrt{\frac{1}{N_c}\sum_{i=1}^{N_c}d_{c,i}^{2}},
\end{equation}
where $N_c$ is the number of evaluated samples in gait cycle $c$. Cycle-wise RMSE values were averaged over all post-calibration cycles to obtain one value per participant. Group results are reported as the mean $\pm$ standard deviation of the five participant-level values. Because phase is normalized by one gait cycle, multiplication by 100 expresses the error as a percentage of the gait cycle (\%GC). For each cycle, the coefficient of determination ($R^2$) was calculated separately for the sine and cosine components, and their arithmetic mean was used as the cycle-level value. Participant- and group-level $R^2$ values were aggregated in the same manner as RMSE.

Given the exploratory cohort size ($n=5$), the analyses were descriptive and no inferential hypothesis tests were performed.

Estimated HS events were identified as resets in the reconstructed predicted phase using a negative-slope threshold, a minimum-amplitude condition, and a temporal refractory period. Each estimated HS was matched one-to-one to an FSR-based event when the absolute temporal difference did not exceed 80~ms. The resulting matches, missed events, and false detections were used to calculate the $F1$-score.

Across the 374 retained gait cycles, the mean cycle duration was $1.45 \pm 0.24$~s, so 80~ms corresponds to approximately 5.5\% of the mean cycle. This criterion therefore evaluates timing near the 5\%GC scale, which is relevant because the timing of phase-scheduled assistance can materially alter biomechanical outcomes~\cite{young2017influence}. For each participant, true positives, false positives, and false negatives were pooled over all post-calibration cycles to obtain one $F1$-score. The five scores were summarized as the group mean $\pm$ standard deviation. Participant-level HS timing error was included only when that participant's $F1$-score was at least 0.5. Group HS timing error was summarized across the included participants, with their number reported, and was marked not interpretable when fewer than two participants met this criterion.

A focused ablation analysis examined the Stage~1 calibration objective and Stage~2 LoRA update policy. Standard fine-tuning (FT) used the same calibration data and $\mathcal{L}_{\mathrm{task}}$ as the primary LwF configuration but omitted the distillation term. Conditional updates were compared with every-cycle and periodic updates every two cycles. Policy-specific hyperparameters were selected within each LOSO fold using only the four training-side participants: Stage~1 selection was performed separately for LwF and FT, followed by Stage~2 selection for each update policy. Each selected configuration was applied to the held-out participant without further tuning. Data order and prequential evaluation were identical across conditions.

\section{Results and Discussion}

\begin{figure}[!t]
    \centering
    \includegraphics[width=\columnwidth]{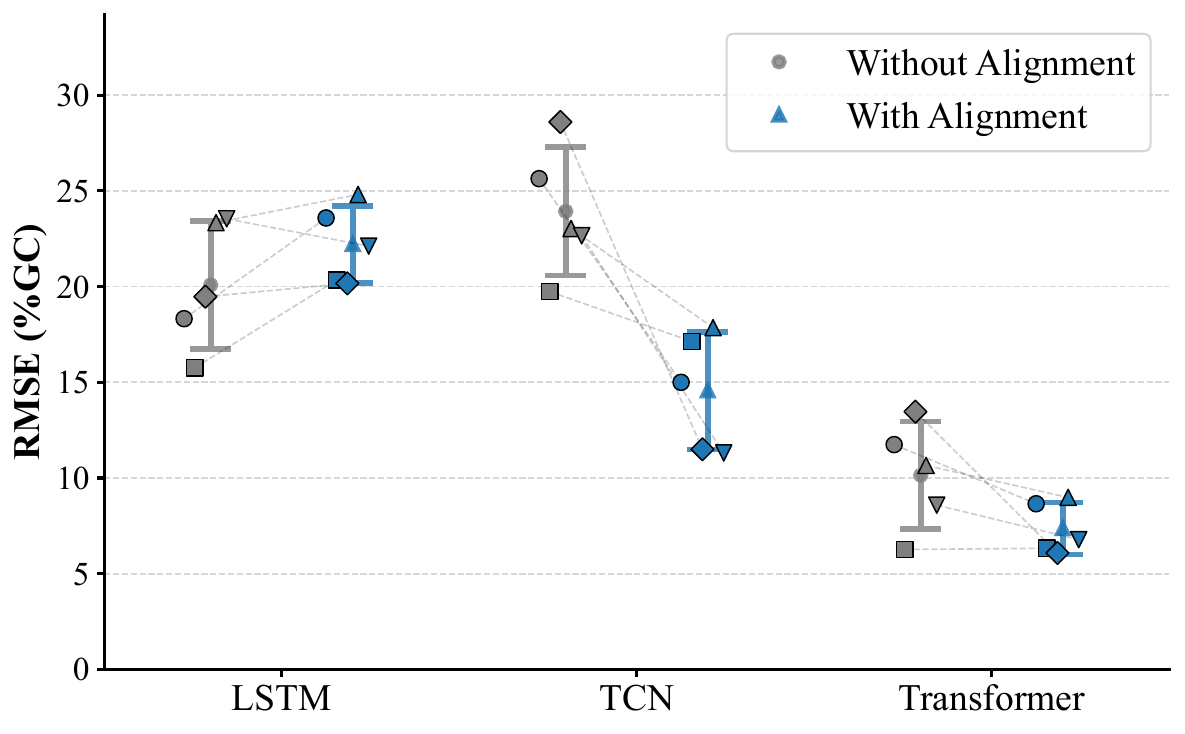}
    \caption{Impact of functional IMU alignment on RMSE (\%GC) across three architectures. Individual participants (S01--S05) are represented by distinct marker shapes ($\circ, \square, \diamond, \vartriangle, \triangledown$). Grey and blue markers denote performance without and with alignment, respectively; group markers and error bars indicate the mean $\pm$ standard deviation across participants.}
    \label{fig:baseline_performance_dot}
\end{figure}

\begin{table}[t]
    \centering
    \caption{Baseline performance comparison with and without functional IMU axis alignment. Values are reported as mean $\pm$ standard deviation across participants.}
    \label{tab:baseline_results}
    \resizebox{\columnwidth}{!}{
    \begin{tabular}{llccc}
    \toprule
    \textbf{Model} & \textbf{Cond.} & $R^2$ & $F1$-score & \shortstack{HS Timing\\Error (ms)} \\
    \midrule
    \multirow{2}{*}{LSTM} & w/o Alig. & $-0.12 \pm 0.25$ & $0.23 \pm 0.28$ & \textit{N/I} \\
     & w/ Alig. & $-0.29 \pm 0.22$ & $0.00 \pm 0.00$ & \textit{N/I} \\
    \midrule
    \multirow{2}{*}{TCN} & w/o Alig. & $-0.49 \pm 0.31$ & $0.00 \pm 0.01$ & \textit{N/I} \\
     & w/ Alig. & $0.23 \pm 0.28$ & $0.09 \pm 0.14$ & \textit{N/I} \\
    \midrule
    \multirow{2}{*}{Trans.} & w/o Alig. & $0.62 \pm 0.17$ & $0.34 \pm 0.27$ & $37.9 \pm 4.8~(n=2)$ \\
     & w/ Alig. & $0.78 \pm 0.07$ & $0.56 \pm 0.26$ & $34.2 \pm 18.0~(n=3)$ \\
    \bottomrule
    \end{tabular}
    }
    \par\vspace{0.3em}
    \parbox{\columnwidth}{\footnotesize HS timing error includes participants with participant-level $F1\geq0.5$; \textit{N/I}: fewer than two participants.}
\end{table}

\subsection{Baseline Model Performance and Impact of IMU Alignment}

We first assessed the zero-shot generalization of the pre-trained models to establish a baseline. Performance differed substantially across architectures and in their responses to functional alignment (Fig.~\ref{fig:baseline_performance_dot} and Table~\ref{tab:baseline_results}). For the LSTM, alignment increased the participant-wise RMSE by 12.3\% on average (group means: $20.09 \pm 3.34$ to $22.20 \pm 2.02$~\%GC; individual changes ranged from a 6.1\% reduction to a 29.1\% increase), while $R^2$ decreased from $-0.12 \pm 0.25$ to $-0.29 \pm 0.22$. Its $F1$-score also decreased from $0.23 \pm 0.28$ to $0.00 \pm 0.00$, indicating that alignment did not improve either continuous phase estimation or HS detection for this architecture.

In contrast, the TCN and Transformer showed improved continuous phase estimation after alignment. The TCN achieved a mean participant-wise RMSE reduction of 37.4\% (range: 13.2--59.8\%), with $R^2$ increasing from $-0.49 \pm 0.31$ to $0.23 \pm 0.28$. However, its aligned $F1$-score remained low ($0.09 \pm 0.14$), indicating that the improvement in continuous phase estimation did not translate into reliable HS detection under the 80~ms matching criterion.

The Transformer achieved a mean participant-wise RMSE reduction of 23.3\% (range: -1.0--54.8\%), and its $R^2$ increased from $0.62 \pm 0.17$ to $0.78 \pm 0.07$. Although S02 showed marginal RMSE degradation, alignment improved the results for the other four participants. The mean $F1$-score increased from $0.34 \pm 0.27$ to $0.56 \pm 0.26$. Among participants with $F1\geq0.5$, HS timing error changed from $37.9 \pm 4.8$~ms ($n=2$) without alignment to $34.2 \pm 18.0$~ms ($n=3$) with alignment. These timing estimates should be interpreted cautiously because they represent subsets of the cohort.

Overall, functional alignment had architecture-dependent effects: it improved continuous phase estimation for the TCN and Transformer but degraded the LSTM, and reliable event detection remained limited at baseline. The aligned Transformer provided the strongest baseline result, but its phase RMSE remained $7.37 \pm 1.35$~\%GC (range: 6.09--8.99~\%GC), and only three participants met the event-detection criterion for interpreting HS timing error. These remaining errors motivated the subsequent evaluation of participant-specific adaptation beyond spatial alignment.

\begin{figure}[t]
    \centering
    \includegraphics[width=\columnwidth]{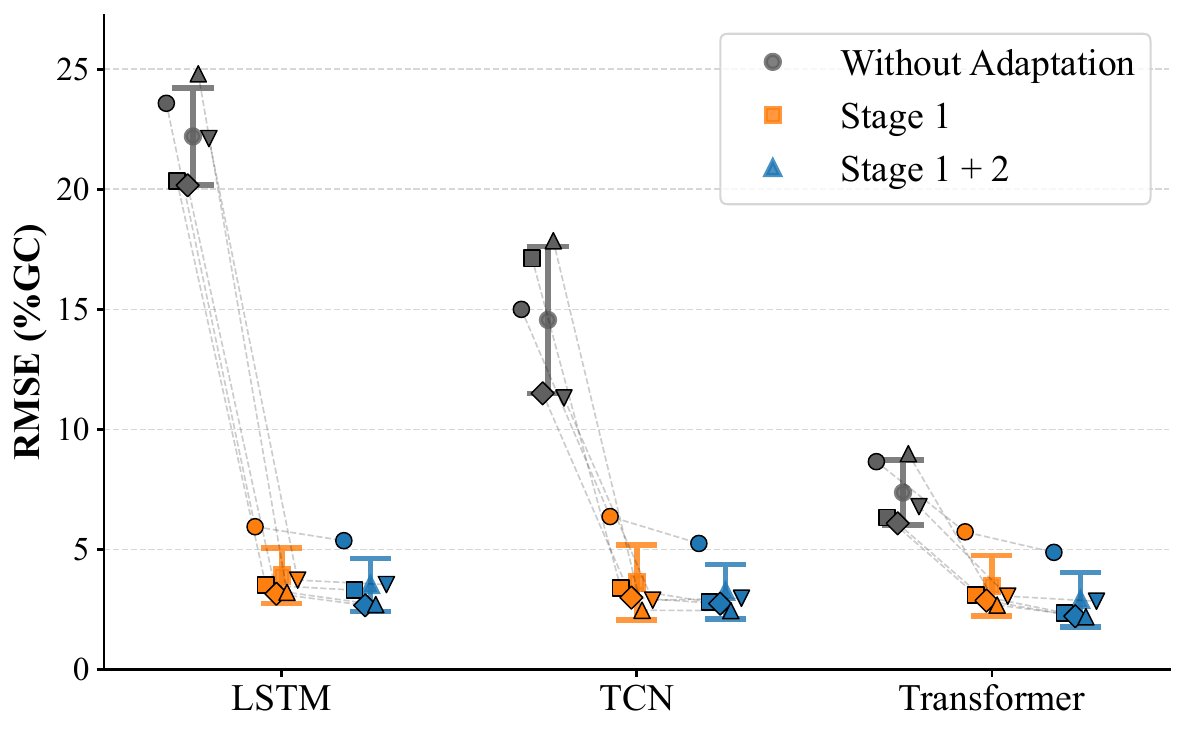}
    \caption{Comparison of phase-estimation RMSE (\%GC) for the baseline, Stage~1, and Stage~1+2 conditions using the LOSO-selected hyperparameters. Individual participants (S01--S05) are distinguished by marker shape ($\circ, \square, \diamond, \vartriangle, \triangledown$); group markers and error bars indicate the mean $\pm$ standard deviation across participants.}
    \label{fig:adaptation_performance_dot}
\end{figure}

\begin{table}[t]
    \centering
    \caption{Performance across adaptation stages using the LOSO-selected hyperparameters. Values are reported as mean $\pm$ standard deviation across participants.}
    \label{tab:personalization_results}
    \resizebox{\columnwidth}{!}{
    \begin{tabular}{llccc}
    \toprule
    \textbf{Model} & \textbf{Cond.} & $R^2$ & $F1$-score & \shortstack{HS Timing\\Error (ms)} \\
    \midrule
    \multirow{3}{*}{LSTM} & w/o Adapt. & $-0.29 \pm 0.22$ & $0.00 \pm 0.00$ & \textit{N/I} \\
     & Stage~1 & $0.92 \pm 0.05$ & $0.80 \pm 0.12$ & $35.1 \pm 3.0$ \\
     & Stage~1+2 & $0.94 \pm 0.04$ & $0.80 \pm 0.12$ & $34.4 \pm 2.8$ \\
    \midrule
    \multirow{3}{*}{TCN} & w/o Adapt. & $0.23 \pm 0.28$ & $0.09 \pm 0.14$ & \textit{N/I}  \\
     & Stage~1 & $0.92 \pm 0.08$ & $0.86 \pm 0.20$ & $33.0 \pm 4.8$ \\
     & Stage~1+2 & $0.94 \pm 0.05$ & $0.87 \pm 0.15$ & $34.5 \pm 4.3$ \\
    \midrule
    \multirow{3}{*}{Trans.} & w/o Adapt. & $0.78 \pm 0.07$ & $0.56 \pm 0.26$ & $34.2 \pm 18.0~(n=3)$ \\
     & Stage~1 & $0.93 \pm 0.06$ & $0.84 \pm 0.19$ & $26.4 \pm 6.8$ \\
     & Stage~1+2 & $0.95 \pm 0.04$ & $0.90 \pm 0.19$ & $23.7 \pm 4.5$ \\
    \bottomrule
    \end{tabular}
    }
    \par\vspace{0.3em}
    \parbox{\columnwidth}{\footnotesize HS timing error includes participants with participant-level $F1\geq0.5$ ($n=5$ unless indicated); \textit{N/I}: fewer than two participants.}
\end{table}

\begin{figure*}
    \centering
    \includegraphics[width=1.0\textwidth]{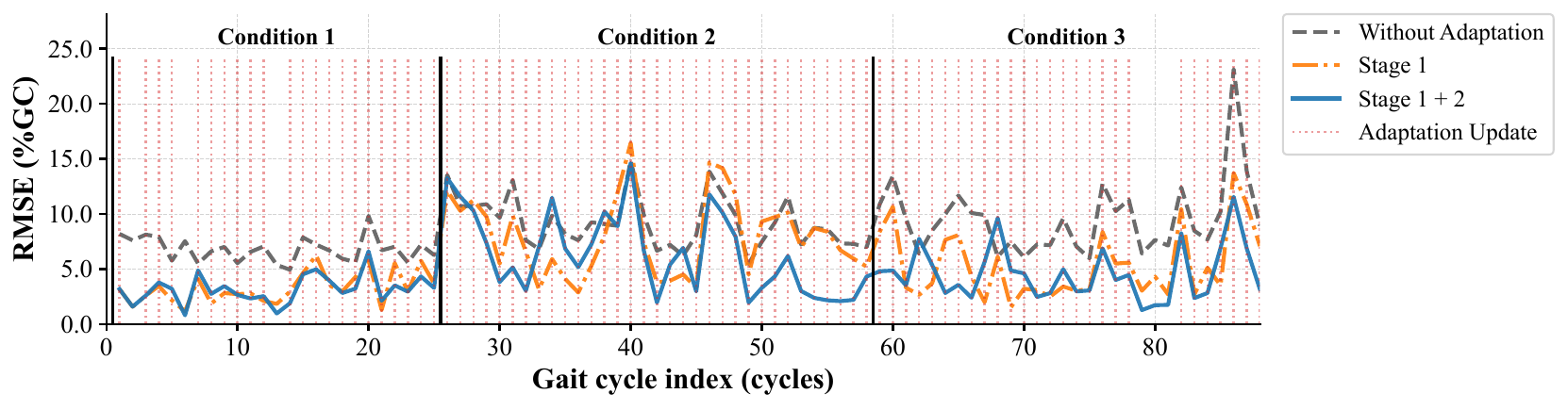}
    \caption{Cycle-wise Transformer RMSE (\%GC) for S01 across adaptation stages and walking conditions using the LOSO-selected hyperparameters. For S01, Conditions 1--3 correspond to NA, PD, and AIF, respectively. Vertical black lines indicate transitions between conditions. Red dotted lines indicate the Stage~2 updates triggered in 83 of 88 evaluated cycles.}
    \label{fig:adaptation_curve}
\end{figure*}

\subsection{Evaluation of the Two-Stage Personalization Strategy}

Stage~1 adaptation produced the largest reduction from the non-adapted baseline across all three architectures (Fig.~\ref{fig:adaptation_performance_dot} and Table~\ref{tab:personalization_results}). The mean RMSE decreased from $22.20 \pm 2.02$ to $3.91 \pm 1.16$~\%GC for the LSTM, from $14.56 \pm 3.07$ to $3.62 \pm 1.57$~\%GC for the TCN, and from $7.37 \pm 1.35$ to $3.49 \pm 1.27$~\%GC for the Transformer. These changes correspond to mean participant-wise reductions of 82.4\% (range: 74.8--87.1\%), 74.5\% (range: 57.5--86.2\%), and 52.5\% (range: 33.8--70.1\%), respectively.

Stage~1 also improved event detection, yielding $F1$-scores of $0.80 \pm 0.12$, $0.86 \pm 0.20$, and $0.84 \pm 0.19$ for the LSTM, TCN, and Transformer, respectively. Participant-specific calibration therefore accounted for most of the improvement from baseline in both continuous phase estimation and HS detection.

Stage~2 provided additional, architecture-dependent refinement. Relative to Stage~1, the mean participant-wise RMSE decreased by 10.3\% for the LSTM (range: 4.6--15.7\%), 8.3\% for the TCN (range: -2.9--17.6\%), and 17.2\% for the Transformer (range: 6.9--23.9\%). The LSTM and Transformer improved in all five participants, whereas the TCN improved in four and showed a 2.9\% increase in S05. After Stage~2, the mean RMSE was $3.52 \pm 1.10$~\%GC for the LSTM, $3.24 \pm 1.14$~\%GC for the TCN, and $2.90 \pm 1.13$~\%GC for the Transformer.

Event-level effects were smaller and model dependent. The LSTM showed a modest increase in $R^2$, while its $F1$-score and HS timing error remained similar. The TCN improved in $R^2$ and maintained a similar $F1$-score, but its mean HS timing error increased from $33.0 \pm 4.8$ to $34.5 \pm 4.3$~ms. The Transformer showed the clearest event-level improvement: $R^2$ increased from $0.93 \pm 0.06$ to $0.95 \pm 0.04$, the $F1$-score increased from $0.84 \pm 0.19$ to $0.90 \pm 0.19$, and HS timing error decreased from $26.4 \pm 6.8$ to $23.7 \pm 4.5$~ms. Thus, Stage~2 primarily refined continuous phase estimation, while its effect on discrete event timing depended on the architecture.

Figure~\ref{fig:adaptation_curve} illustrates the cycle-wise Transformer results for S01, the participant with the highest mean error after adaptation. The Stage~1+2 RMSE averaged 3.15~\%GC under no assistance and increased to 6.43 and 4.59~\%GC under the PD and AIF conditions, respectively. Relative to Stage~1, Stage~2 reduced the overall S01 RMSE by 15.1\%. The number of cycles exceeding 10~\%GC decreased from 13 to 9, and the maximum cycle-wise error decreased from 16.44 to 14.56~\%GC. The conditional criterion triggered 83 updates over 88 evaluated cycles, reflecting frequent deviations from the Stage~1 reference loss during the assisted conditions. Thus, conditional updating should not be interpreted as uniformly sparse; its update frequency varied with the observed cycle-level error. Cycle-wise results for the remaining participants and architectures are provided in Fig.~S1--S5.

Overall, the LOSO-selected two-stage strategy yielded mean participant-wise RMSE reductions of 84.2\% for the LSTM, 77.0\% for the TCN, and 60.7\% for the Transformer relative to the non-adapted baseline. Among the evaluated architectures, the Transformer achieved the lowest final mean RMSE ($2.90 \pm 1.13$~\%GC). Although the adaptation and evaluation protocols differed, this mean error was in the same numerical range as the $2.9 \pm 0.6$~\%GC reported by Kang \textit{et al.} for rapid online adaptation in stroke survivors~\cite{kang2025online}. The larger between-participant variability in the present study, however, indicates greater participant-dependent variation in the resulting phase-estimation accuracy. Because the initial calibration condition differed across participants, participant-specific and calibration-condition effects could not be separated; future studies should standardize or counterbalance this condition.

The present study evaluated adaptation within a single recording session as gait cycles progressed across successive trials and assistance conditions. Thus, the results characterize cycle- and trial-scale refinement but do not establish performance across days or rehabilitation sessions. Longer-term use may introduce sensor-remounting variation, changes in fatigue or impairment, and cumulative parameter drift from repeated updates. Sequential updates could lead to forgetting if adaptation to a newly encountered assistance condition degrades performance when the user returns to a previously encountered condition. Because the present protocol did not revisit earlier conditions after prolonged adaptation, and because walking condition and elapsed time changed together, it could not assess such backward retention. Repeated-session studies should evaluate both adaptation to new conditions and retention on previous conditions, together with criteria for functional re-alignment, Stage~1 recalibration, or model reset. In addition, the small cohort consisted only of men with ischemic stroke and FAC levels 3--4, limiting generalization to women, hemorrhagic stroke, and individuals with more severe walking impairment.

\subsection{Effect of Adaptation Design Choices}

\begin{table}[t]
    \centering
    \caption{Ablation analysis of the Stage~1 objective and Stage~2 LoRA update policy using policy-specific LOSO-selected hyperparameters. Values are reported as mean $\pm$ standard deviation across participants.}
    \label{tab:ablation_design_choices}
    \resizebox{\columnwidth}{!}{
    \begin{tabular}{lllcc}
    \toprule
    \textbf{Model} & \textbf{Stage~1} & \textbf{Stage~2 LoRA policy} & \textbf{RMSE (\%GC)} & \textbf{Updates} \\
    \midrule
    \multirow{8}{*}{LSTM}
     & FT & None (Stage~1 only) & $3.68 \pm 1.41$ & $0$ \\
     & FT & Conditional & $3.47 \pm 1.54$ & $43.8 \pm 9.7$ \\
     & FT & Every cycle & $3.45 \pm 1.54$ & $69.4 \pm 14.1$ \\
     & FT & Periodic (every 2 cycles) & $3.53 \pm 1.52$ & $34.6 \pm 7.0$ \\
     & LwF & None (Stage~1 only) & $3.91 \pm 1.16$ & $0$ \\
     & LwF & Conditional & $3.52 \pm 1.10$ & $48.0 \pm 21.9$ \\
     & LwF & Every cycle & $3.41 \pm 1.12$ & $69.4 \pm 14.1$ \\
     & LwF & Periodic (every 2 cycles) & $3.55 \pm 1.12$ & $34.6 \pm 7.0$ \\
    \midrule
    \multirow{8}{*}{TCN}
     & FT & None (Stage~1 only) & $3.59 \pm 1.93$ & $0$ \\
     & FT & Conditional & $3.19 \pm 1.33$ & $57.2 \pm 18.9$ \\
     & FT & Every cycle & $3.18 \pm 1.34$ & $69.4 \pm 14.1$ \\
     & FT & Periodic (every 2 cycles) & $3.32 \pm 1.48$ & $34.6 \pm 7.0$ \\
     & LwF & None (Stage~1 only) & $3.62 \pm 1.57$ & $0$ \\
     & LwF & Conditional & $3.24 \pm 1.14$ & $52.4 \pm 20.5$ \\
     & LwF & Every cycle & $3.22 \pm 1.15$ & $69.4 \pm 14.1$ \\
     & LwF & Periodic (every 2 cycles) & $3.38 \pm 1.23$ & $34.6 \pm 7.0$ \\
    \midrule
    \multirow{8}{*}{Trans.}
     & FT & None (Stage~1 only) & $3.41 \pm 1.55$ & $0$ \\
     & FT & Conditional & $2.85 \pm 1.21$ & $63.0 \pm 13.9$ \\
     & FT & Every cycle & $2.83 \pm 1.22$ & $69.4 \pm 14.1$ \\
     & FT & Periodic (every 2 cycles) & $2.93 \pm 1.24$ & $34.6 \pm 7.0$ \\
     & LwF & None (Stage~1 only) & $3.49 \pm 1.27$ & $0$ \\
     & LwF & Conditional & $2.90 \pm 1.13$ & $53.8 \pm 19.2$ \\
     & LwF & Every cycle & $2.86 \pm 1.13$ & $69.4 \pm 14.1$ \\
     & LwF & Periodic (every 2 cycles) & $2.94 \pm 1.18$ & $34.6 \pm 7.0$ \\
    \bottomrule
    \end{tabular}
    }
\end{table}

Table~\ref{tab:ablation_design_choices} summarizes the policy-specific LOSO ablation of the Stage~1 objective and Stage~2 update policy. Participant-level results are shown in Fig.~S6. Adding Stage~2 updates reduced the mean RMSE relative to the corresponding Stage~1-only condition across all architectures. Standard fine-tuning yielded slightly lower mean RMSE than LwF for all three architectures in both the Stage~1-only and conditional-update comparisons. At the participant level, fine-tuning had lower RMSE in 9 of 15 Stage~1-only comparisons and 11 of 15 conditional-update comparisons. The paired differences (LwF minus fine-tuning) ranged from $-0.65$ to $0.68$~\%GC and from $-0.84$ to $0.63$~\%GC, respectively, showing that the direction was not uniform across participants.

Across both Stage~1 objectives, every-cycle updating yielded the lowest group-mean RMSE for all three architectures, followed by conditional and periodic updating; each Stage~2 policy also improved upon its corresponding Stage~1-only condition. Under conditional updating, the group-mean difference between the FT and LwF configurations was $0.05$~\%GC for each architecture. Thus, the present results do not demonstrate an immediate accuracy advantage of LwF, but the qualitative comparison of the Stage~2 update policies was unchanged across the two calibration objectives.

LwF was adopted as a conservative Stage~1 regularizer because participant-specific calibration relied on a single short walking trial. Its distillation term constrains departures from the pre-trained mapping while the FSR-supervised task loss drives personalization. All 15 primary LOSO folds selected $\lambda=0.1$, the smallest candidate value. Within the evaluated grid, this indicates that stronger distillation was not favored under direct FSR supervision and within-session evaluation. Because smaller nonzero weights were not evaluated, this boundary selection does not establish the optimal regularization strength.

Accordingly, LwF should be interpreted as a conservative regularized calibration choice rather than as an empirically superior component of the framework. Standard fine-tuning remains a simpler and competitive alternative when short-term personalization accuracy is the primary objective. Evaluation across different calibration durations and levels of supervision quality would be needed to determine when distillation regularization provides a measurable benefit.

Despite the small accuracy advantage of every-cycle updating, the conditional policy was retained as the primary Stage~2 strategy because it provided an error-responsive compromise between every-cycle and fixed-period updating. In the LwF-based configuration, it reduced the mean update count relative to every-cycle LoRA by 30.8\%, 24.5\%, and 22.5\% for the LSTM, TCN, and Transformer, respectively, while increasing RMSE by only 0.11, 0.02, and 0.04~\%GC. Periodic updating further reduced the update count but produced higher error in most comparisons and did not respond to the observed cycle loss.

The conditional trigger did not enforce sparse updates in every participant because its frequency depended on how often the cycle loss exceeded the Stage~1 reference loss. In the primary LOSO selection, all 15 folds selected the largest evaluated LoRA rank ($r=16$), and 13 selected the smallest trigger threshold ($\gamma=1$). These boundary selections are consistent with the preference for greater update capacity and the low error of every-cycle updating, and the selected values should therefore be interpreted within the evaluated candidate ranges. Overall, conditional Stage~2 provides error-responsive low-rank refinement with an accuracy--update trade-off that depends on the selected model capacity and trigger threshold.

\begin{figure}[t]
    \centering
    \includegraphics[width=\columnwidth]{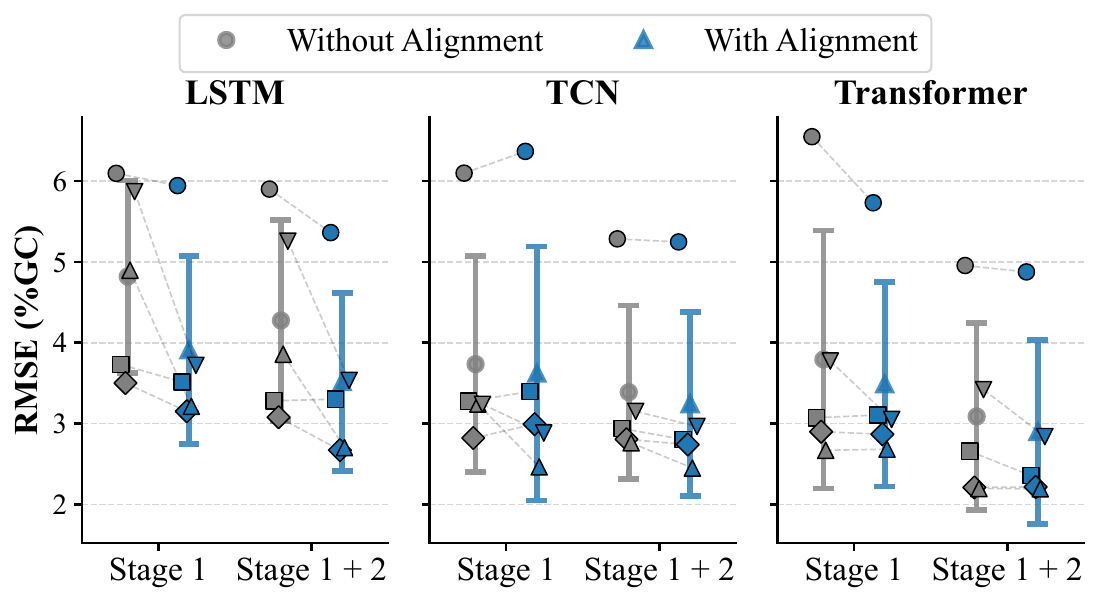}
    \caption{Effect of functional IMU alignment after Stage~1 and Stage~1+2 personalization using the LOSO-selected hyperparameters. Individual participants (S01--S05) are distinguished by marker shape ($\circ, \square, \diamond, \vartriangle, \triangledown$); group markers and error bars indicate the mean $\pm$ standard deviation across participants.}
    \label{fig:imu_alignment_personalization}
\end{figure}

\begin{table}
    \caption{Performance comparison of the final personalized models with and without functional IMU alignment using the LOSO-selected hyperparameters. Values are reported as mean $\pm$ standard deviation across participants.}
    \label{tab:adaptation_alignment_results}
    \resizebox{\columnwidth}{!}{
    \begin{tabular}{llccc}
    \toprule
    \textbf{Model} & \textbf{Cond.} & $R^2$ & $F1$-score & \shortstack{HS Timing\\Error (ms)} \\
    \midrule
    \multirow{2}{*}{LSTM} & Adapt. w/o Alig. & $0.91 \pm 0.05$ & $0.79 \pm 0.16$ & $37.2 \pm 5.1$ \\
     & Adapt. w/ Alig. & $0.94 \pm 0.04$ & $0.80 \pm 0.12$ & $34.4 \pm 2.8$ \\
    \midrule
    \multirow{2}{*}{TCN} & Adapt. w/o Alig. & $0.94 \pm 0.04$ & $0.84 \pm 0.15$ & $34.0 \pm 5.7$ \\
     & Adapt. w/ Alig. & $0.94 \pm 0.05$ & $0.87 \pm 0.15$ & $34.5 \pm 4.3$ \\
    \midrule
    \multirow{2}{*}{Trans.} & Adapt. w/o Alig. & $0.95 \pm 0.04$ & $0.89 \pm 0.16$ & $31.0 \pm 5.7$ \\
     & Adapt. w/ Alig. & $0.95 \pm 0.04$ & $0.90 \pm 0.19$ & $23.7 \pm 4.5$ \\
    \bottomrule
    \addlinespace[2pt]
    \end{tabular}
    }
\end{table}

\begin{table*}[t]
\centering
\caption{Concurrent 100-Hz inference and Stage~2 adaptation timing. Inference values are summarized across participants, and adaptation values across triggered updates.}
\label{tab:realtime_feasibility}
\begin{tabular}{lcccccc}
\toprule
\multirow{2}{*}{\textbf{Model}} &
\multicolumn{3}{c}{\textbf{Inference}} &
\multicolumn{3}{c}{\textbf{Adaptation Time (ms)}} \\
\cmidrule(lr){2-4}\cmidrule(lr){5-7}
& \textbf{Mean (ms)} & \textbf{p99 (ms)} & \textbf{Miss (\%)}
& \textbf{Mean} & \textbf{p99} & \textbf{Maximum} \\
\midrule
LSTM & $5.3 \pm 0.5$ & $29.6 \pm 3.6$ & $6.6$ & $244.4 \pm 50.3$ & $358.9$ & $413.2$ \\
TCN & $0.7 \pm 0.1$ & $3.3 \pm 0.3$ & $0.0$ & $182.1 \pm 33.0$ & $268.6$ & $280.2$ \\
Transformer & $0.9 \pm 0.1$ & $3.2 \pm 0.3$ & $0.0$ & $393.4 \pm 74.9$ & $550.9$ & $589.6$ \\
\bottomrule
\end{tabular}
\end{table*}

\subsection{Effect of IMU Alignment on Personalization}

We examined whether functional IMU alignment provided additional benefits after personalization with LOSO-selected hyperparameters. After Stage~1+2, alignment changed the mean RMSE from $4.28 \pm 1.25$ to $3.52 \pm 1.10$~\%GC for the LSTM, from $3.39 \pm 1.07$ to $3.24 \pm 1.14$~\%GC for the TCN, and from $3.09 \pm 1.16$ to $2.90 \pm 1.13$~\%GC for the Transformer. The corresponding mean participant-wise reductions were 16.8\% (range: -0.8--32.5\%), 5.0\% (range: 0.7--11.6\%), and 5.7\% (range: -1.6--16.9\%), respectively. Alignment improved all five TCN participants and four participants for each of the LSTM and Transformer. Thus, its RMSE effect was largest for the LSTM but was not uniform across participants.

Event-level results showed a more consistent benefit in HS detection (Table~\ref{tab:adaptation_alignment_results}). Alignment preserved or increased the mean $F1$-score across the architectures and reduced HS timing error from $37.2 \pm 5.1$ to $34.4 \pm 2.8$~ms for the LSTM and from $31.0 \pm 5.7$ to $23.7 \pm 4.5$~ms for the Transformer. The TCN $F1$-score increased from $0.84 \pm 0.15$ to $0.87 \pm 0.15$, while its timing error remained comparable.

These results suggest that the complementary value of functional alignment after personalization may lie primarily in HS detection and timing rather than in a uniform reduction in cycle-averaged phase error. Relative to the mean gait-cycle duration of 1.45~s, the final HS timing errors corresponded to approximately 2.4, 2.4, and 1.6~\%GC for the LSTM, TCN, and Transformer, respectively. Prior exoskeleton studies have shown that assistance timing can affect delivered mechanical power, metabolic cost, and gait parameters~\cite{ding2016effect, lee2017effects}. These values characterize estimator-side event timing only; closed-loop suitability cannot be inferred without evaluating the controller, actuator dynamics, and resulting biomechanical response.

\subsection{Embedded Timing of Concurrent Inference and Adaptation}

Concurrent inference and Stage~2 adaptation were evaluated on the embedded platform by replaying data from all five participants at the original 100-Hz sampling rate. All triggered updates were completed and applied without skipped updates or adaptation backlog.

As shown in Table~\ref{tab:realtime_feasibility}, the TCN and Transformer completed all Stage~2 inferences within the 10-ms sampling interval. Their participant-level p99 latencies were $3.3 \pm 0.3$ and $3.2 \pm 0.3$~ms, respectively. In contrast, although the LSTM achieved a mean latency below 10~ms, its p99 latency increased to $29.6 \pm 3.6$~ms and 6.6\% of its inferences exceeded the deadline. The violations occurred primarily when inference overlapped with adaptation, indicating architecture-dependent contention on the shared embedded graphics processor.

The adaptation p99 times were $358.9$, $268.6$, and $550.9$~ms for the LSTM, TCN, and Transformer, respectively. Their corresponding maxima were $413.2$, $280.2$, and $589.6$~ms. Every measured update therefore completed within 0.8~s, shorter than the gait-cycle duration reported for slow walking ($>0.8$~s)~\cite{winter2009biomechanics}.

Because updates were executed asynchronously and applied at subsequent cycle boundaries, the TCN and Transformer results support the timing feasibility of concurrent inference and adaptation under the evaluated replay conditions. The LSTM requires further scheduling or computational optimization for concurrent 100-Hz operation. This evaluation used offline replay; validation during online closed-loop exoskeleton control remains future work.

\section{Conclusion}

This study evaluated functional IMU alignment and two-stage sequential adaptation for personalized gait-phase estimation in stroke survivors. The estimator used thigh-mounted IMU signals, with heel-FSR-derived phase labels supporting offline adaptation and evaluation.

Relative to non-adapted baselines, Stage~1+2 reduced mean participant-wise RMSE by 84.2\%, 77.0\%, and 60.7\% for the LSTM, TCN, and Transformer, respectively. The Transformer achieved the lowest final RMSE ($2.90 \pm 1.13$~\%GC) and HS timing error ($23.7 \pm 4.5$~ms). Functional alignment produced model-dependent RMSE changes while preserving or improving HS detection. Ablation results showed no accuracy advantage of LwF over standard fine-tuning, whereas Stage~2 improved upon the corresponding Stage~1-only models. Every-cycle updating generally minimized error, while conditional LoRA reduced updates with small accuracy differences. Together, these findings identify low-rank post-calibration refinement and its update policy as central design elements of the two-stage framework.

In embedded replay, the TCN and Transformer maintained concurrent 100-Hz inference without deadline misses, whereas the LSTM missed 6.6\% of deadlines. All updates completed within 0.8~s without backlog. These results support computational timing feasibility under the evaluated replay conditions.

The study was limited to within-session offline replay, and long-term retention across independent sessions remains to be evaluated. Future work should replace FSR-derived supervision with IMU-based gait-event detection or thigh-kinematics landmarks~\cite{ye2020adaptive, villarreal2017robust}, and evaluate larger cohorts, repeated sessions, and online closed-loop integration.

\section*{Data and Code Availability}
The dataset and training code for the two-stage personalization procedure are publicly available at \href{https://github.com/hyungseok-ryu/personalized-gait-phase-estimation}{\underline{github.com/hyungseok-ryu/personalized-gait-phase-estimation}}.

\bibliographystyle{IEEEtran} 
\bibliography{reference}

\end{document}